\documentclass[final]{siamart1116}

\usepackage{lipsum}
\usepackage{amsfonts}
\usepackage{graphicx}
\usepackage{epstopdf}
\usepackage{algorithmic}
\ifpdf
  \DeclareGraphicsExtensions{.eps,.pdf,.png,.jpg}
\else
  \DeclareGraphicsExtensions{.eps}
\fi

\usepackage{xspace}

\newcommand{\MyExp}[2]{\ensuremath{\mathbf{Exp}_{#1}(#2)}\xspace}
\newcommand{\MyLog}[2]{\ensuremath{\mathbf{Log}_{#1}(#2)}\xspace}
\newcommand{\Rn}[1]{\ensuremath{\mathbb{R}^{#1}}\xspace}
\newcommand{\Sn}[1]{\ensuremath{\mathcal{S}^{#1}}\xspace}

\newcommand{\TSnm}[1]{\ensuremath{\mathcal{T}_{#1}\mathcal{S}^{n-1}}\xspace}

\numberwithin{theorem}{section}

\newcommand{\TheTitle}{Tight Semi-Nonnegative Matrix Factorization} 
\newcommand{\TheAuthors}{David W. Dreisigmeyer}

\headers{\TheTitle}{\TheAuthors}

\title{{\TheTitle}\thanks{
Any opinions and conclusions expressed herein are those of the author and do not necessarily represent the views of the U.S. Census Bureau.
The research in this paper does not use any confidential Census Bureau information.
This was authored by an employee of the US national government. 
As such, the Government retains a nonexclusive, royalty-free right to publish or reproduce this article, or to allow others to do so, for Government purposes only.
}}

\author{
	David W. Dreisigmeyer\thanks{United States Census Bureau, 
Center for Economic Studies, 
Suitland, MD and
Department of Electrical and Computer Engineering, 
Colorado State University, 
Fort Collins, CO
    (\email{david.wayne.dreisigmeyer@census.gov})}
}

\usepackage{amsopn}

\ifpdf
\hypersetup{
  pdftitle={\TheTitle},
  pdfauthor={\TheAuthors}
}
\fi

\begin{document}

\maketitle

\begin{abstract}
  The nonnegative matrix factorization is a widely used, flexible matrix decomposition, finding applications in biology, image and signal processing and information retrieval, among other areas.
  Here we present a related matrix factorization.
  A multi-objective optimization problem finds conical combinations of templates that approximate a given data matrix.
  The templates are chosen so that as far as possible only the initial data set can be represented this way.
  However, the templates are not required to be nonnegative nor convex combinations of the original data.
\end{abstract}



\section{\label{sec:introduction}Introduction}
The nonnegative matrix factorization (NMF) is a popular and useful matrix decomposition.
It has been employed in image processing \cite{art:lee-1999}, clustering \cite{art:ding-2005}, biology \cite{art:brunet-2004} and information retrieval \cite{art:ding2008}. 
A data matrix $X$, $x_{ij} \geq 0$, is approximated by
\begin{equation*}
\label{eq:nmf}
X \approx WH
\end{equation*}
where $W$ and $H$ both have only nonnegative entries, denoted by $0 \leq W$ and $0 \leq H$.
With $X \in \Rn{n \times m}$ and an \textit{a priori} chosen $k \in \mathbb{N}$, then $W \in \Rn{n \times k}$ and $H \in \Rn{k \times m}$.
There are multiple extensions of the NMF, see, e.g., \cite{art:zhang-2013}.
Two such extensions are to only require $H$ to be non-negative and/or to require that the columns of $W$ be convex combinations of the original data.
These are called the semi-nonnegative matrix factorization (SNMF) and convex matrix factorization (CMF), respectively \cite{art:ding-2010}.

Here we look at having the matrix decomposition be the result of a multi-objective optimization problem.
There will be two objectives to be satisfied: 
1) minimizing the error in the approximation of the original data (model fidelity), and 
2) minimizing the maximum angle between any two columns of $W$ (model tightness).
This method is closely related to SNMF and CMF as well as archetypal analysis (AA) \cite{art:cutler-1994, art:damle-2016, misc:javadi-2017}.
However, we do not require that the templates, i.e., the columns of $W$, be convex combinations of the original data in $X$.
A volume minimization criterion has been included in the NMF previously to find tight models \cite{art:chan-2009, art:fu-2016, art:lin-2015, art:miao-2007, art:zhou-2011}.
While related to our methodology, volume minimization is not the correct measure of model tightness to use in our new method.

The paper is organized as follows.  
In Section~\ref{sec:algorithm} the new algorithm is presented.  
Section~\ref{sec:solution} presents a solution method to the factorization problem.
An example is looked at in Section~\ref{sec:examples}.
Finally, a discussion follows in Section~\ref{sec:discussion}.

\section{\label{sec:algorithm}Tight Semi-nonnegative Matrix Factorization}
The regular NMF for a nonnegative matrix $X \in \Rn{n \times m}$ solves the optimization problem
\begin{eqnarray*}
\min_{0 \leq W,0 \leq H} & & d(X,WH) \label{eq:reg_nmf_alg} 
\end{eqnarray*}
where $W \in \Rn{n \times k}$ and $H \in \Rn{k \times m}$ for a chosen $k \in \mathbb{Z}$, and $d(X,WH)$ is a measure of distance or similarity between $X$ and $WH$.
Typically $d(X,WH)$ will be the squared Frobenius norm \cite{art:lee-2000}
\begin{eqnarray*}
\label{eq:frob_norm}
d(X,WH) 
	& = & \sum_{ij} \left( X - WH \right)^{2}_{ij} \\
    & = & \|X - WH\|_{F}^{2} \nonumber
\end{eqnarray*}
or a modified Kullback-Leibler divergence
\begin{eqnarray*}
\label{eq:kl_div}
d(X,WH) 
	& = & \sum_{ij} X_{ij} \log \frac{X_{ij}}{(WH)_{ij}} - X_{ij} + (WH)_{ij} \mbox{.}
\end{eqnarray*}
We let $d(X,WH) = \|X - WH\|_{F}^{2}$ in the remainder of the paper.

There are many extensions to the NMF and we now look at a few of them that are most relevant.
For the SNMF the conditions on $X$ and $W$ are removed so that we have
\begin{eqnarray}
\min_{W,0 \leq H} & & d(X,WH) \mbox{.} \label{eq:snmf_alg}
\end{eqnarray}
For CMF the templates are convex combinations of the original data.
Then the optimization problem is stated as
\begin{eqnarray*}
\min_{0 \leq W,0 \leq H} & & d(X,XWH) \label{eq:cmf_alg-1} \\
\mathrm{subject\ to\ } 
	& & W^{T} \mathbf{1} = \mathbf{1} \label{eq:cmf_alg-2b} \mbox{.}
\end{eqnarray*}
Finally, the extended version of AA additionally requires that $H$ be column stochastic which results in
\begin{eqnarray*}
\min_{0 \leq W,0 \leq H} & & d(X,XWH) \label{eq:aa_alg-1} \\
\mathrm{subject\ to\ } 
	& & W^{T} \mathbf{1} = \mathbf{1} \label{eq:aa_alg-2b} \\
	& & H^{T} \mathbf{1} = \mathbf{1} \label{eq:aa_alg-2d} \mbox{.}
\end{eqnarray*}

$W$ defines the hyper-area $A(W)$ of a $k$-dimensional parallelogram.
With the singular value decomposition $W = U S V^{T}$ we have that
\begin{eqnarray*}
\label{eq:W_area}
A(W) 
	& \doteq & \prod_{i} s_{ii} \mbox{.}
\end{eqnarray*}
Including area minimization in a NMF has been looked at previously \cite{art:chan-2009, art:fu-2016, art:lin-2015, art:miao-2007, art:zhou-2011}.
In \cite{art:miao-2007} the optimization problem is
\begin{subequations}
\label{eq:miao_alg}
\begin{eqnarray}
\min_{0 \leq W,0 \leq H} & & d(X,WH) + \lambda A(W) \label{eq:miao_alg-1} \\
\mathrm{subject\ to\ } 
	& & H^{T} \mathbf{1} = \mathbf{1} \label{eq:miao_alg-2}
\end{eqnarray}
\end{subequations}
where $\lambda > 0$.
The requirement that $W$ be non-negative can be removed.
The equality constraints on the columns of $H$ can be replaced by equality constraints on the columns of $W$:
\begin{subequations}
\label{eq:zhou_alg}
\begin{eqnarray}
\min_{0 \leq W,0 \leq H} & & d(X,WH) + \lambda A(W) \label{eq:zhou_alg-1} \\
\mathrm{subject\ to\ } 
	& & W^{T} \mathbf{1} = \mathbf{1} \label{eq:zhou_alg-2}
\end{eqnarray}
\end{subequations}
which is similar to the method presented in \cite{art:zhou-2011}.
Here the non-negativity of $W$ is needed.
Notice that (\ref{eq:miao_alg}) requires that the approximation to $X$ be given by convex combinations of the columns of $W$ while (\ref{eq:zhou_alg}) only requires that the data lie within the cone defined by the columns of $W$.

Both (\ref{eq:miao_alg}) and (\ref{eq:zhou_alg}) have the obvious restatements as standard multi-objective optimization (MOO) problems.
With the vector-valued objective function $\mathbf{f}_{X}(W,H) = [ d(X,WH)\ A(W) ]$ the MOO problem to solve is
\begin{eqnarray*}
\min_{0 \leq W,0 \leq H} & & \mathbf{f}_{X}(W,H) = \left[ d(X,WH)\ A(W) \right]\label{eq:moo_alg-1} \\
\mathrm{subject\ to\ } 
	& & H^{T} \mathbf{1} = \mathbf{1} \mbox{ or } W^{T} \mathbf{1} = \mathbf{1} \label{eq:moo_alg-2} \mbox{.}
\end{eqnarray*}
Forming the scalar objective function $[1\ \lambda] \cdot \mathbf{f}_{X}(W,H)$ the Pareto frontier can be constructed by using different values for $\lambda \geq 0$.
Each point on the Pareto frontier gives the optimal trade-off between model fidelity, given by the $d(X,WH)$ term, and model `tightness', given by the $A(W)$ term.
The columns of $W$ can be rescaled if the rows of $H$ are inversely scaled, so a normalization condition needs to be placed on either $W$ or $H$ otherwise $A(W)$ can be made arbitrarily small.

We wish to generalize the SNMF so any normalization of a column of $A$ should be independent of the direction of the column.
This suggests requiring $\| \mathbf{w} \|_{2} = 1$ for each column of $W$.
An underlying assumption is that the data points lie within, rather than outside of, the cone defined by conical combinations of $W$'s columns.
By this we mean that all of the data points lie within an open half-sphere of the unit hypersphere \Sn{n-1} after the columns of $X$ have been normalized to unit $2$-norm.
In particular, this guarantees the existence and uniqueness of the Karcher mean for the data \cite{proc:krakowski-2007}.
But this also means that minimizing $A(W)$ is not the correct measure of tightness.
As an example, let $W \in \Rn{n \times 2}$ and consider $A(W) = \sqrt{\det(W^T W)}$ where
\begin{equation*}
W^{T} W = 
			\left[
            	\begin{array}{cc}
            	1 & \pm a \\
                \pm a & 1
            	\end{array}
			\right] \mbox{.}
\end{equation*}
We see that $A(W) = \sqrt{1 - a^2}$ is the same when the angle between $\mathbf{w}_{1}$ and $\mathbf{w}_{2}$ is $\theta$ or $\pi - \theta$ for $0 < \theta < \pi/2$.
Since we allow the normalized data points to lie in an open half-sphere there's no \textit{a priori} reason to restrict the angles between the columns of $W$ to be acute or obtuse.

A better measure of model tightness is the maximum geodesic distance
\begin{equation}
\label{eq:geo-dist}
S(W) = \max_{i < j} \arccos( \mathbf{w}_i \cdot \mathbf{w}_j )
\end{equation}
over \Sn{n-1} between any two columns of $W$.
Now we have the MOO problem
\begin{subequations}
\label{eq:moo_nmf_alg}
\begin{eqnarray}
\min_{W,0 \leq H} & & \mathbf{f}_{X}(W,H) = \left[ d(X,WH)\ S(W) \right]\label{eq:moo_nmf_alg-1} \\
\mathrm{subject\ to\ } 
	& & \| \mathbf{w}_{i} \|_{2} = 1 \label{eq:moo_alg-2b}
\end{eqnarray}
\end{subequations}
where $W = [\mathbf{w}_{1} | \cdots | \mathbf{w}_{k} ]$.
This is the tight semi-nonnegative matrix factorization (tSNMF) of $X$.
Since $S(W)$ is as small as possible while still giving the same approximation of the data, $W$ models the original data $X$ but as little else as possible.
The available solutions are given by the Pareto frontier defined by (\ref{eq:moo_nmf_alg}).

\section{\label{sec:solution}Solution Method}
The MOO problem in (\ref{eq:moo_nmf_alg}) can be restated by placing an upper bound on $S(W)$ and treating this as an inequality constraint.
Now (\ref{eq:moo_nmf_alg}) is restated as
\begin{eqnarray*}
\min_{W,0 \leq H} & & \|X - WH\|_{F}^{2} \label{eq:tsnmf_alg-1} \\
\mathrm{subject\ to\ }
	& &	S(W) \leq \epsilon \label{eq:tsnmf_alg-2a} \\
    & & \| \mathbf{w}_{i} \|_{2} = 1 \label{eq:tsnmf_alg-2b}
\end{eqnarray*}
where different values of $\epsilon$ give different points on the Pareto frontier.
(This is an alternate way of working with MOO problems versus forming the scalar objective function $[1\ \lambda] \cdot \mathbf{f}_{X}(W,H)$.)
Because of the equality constraints the optimization problem occurs on a Riemannian manifold \cite{misc:dreisigmeyer-2006}.
This is an area of optimization that has proven very fruitful over the last few decades.
See, for example, the seminal article by Edelman, Arias and Smith \cite{art:edelman-1999}.

Maintaining the vector norm equality constraints is accomplished by restricting the optimization to occur over the unit hypersphere \Sn{n - 1}.
A reasonable way of doing this is to first find the Karcher mean $\bar{\mathbf{x}}$ of the normalized data $\widehat{X}$, where the individual data point $\hat{\mathbf{x}}_{i}$ is the original data point $\mathbf{x}_{i}$ normalized to unit length.
We work strictly in the tangent space \cite{proc:krakowski-2007}
\begin{equation*}
\TSnm{\bar{\mathbf{x}}} = \left\lbrace \mathbf{v} \ | \ \bar{\mathbf{x}} \cdot \mathbf{v} = 0  \right\rbrace
\end{equation*}
which is sufficient since we assumed that the unit-normalized data points lie within an open half-sphere of \Sn{n-1}.
A vector $\mathbf{v} \in \TSnm{\bar{\mathbf{x}}}$ is mapped onto \Sn{n-1} with the exponential map
\begin{equation*}
\MyExp{\bar{\mathbf{x}}}{\mathbf{v}} 
	= \cos\|\mathbf{v}\| \bar{\mathbf{x}} + \sin\|\mathbf{v}\| \frac{\mathbf{v}}{\|\mathbf{v}\|} \mbox{,}
\end{equation*}
$\| \mathbf{v} \| < \pi / 2$.
The inverse mapping from \Sn{n-1} to $\TSnm{\bar{\mathbf{x}}}$ is
\begin{equation*}
\MyLog{\bar{\mathbf{x}}}{\mathbf{w}} 
	= \frac
    	{\arccos(\bar{\mathbf{x}} \cdot \mathbf{w})}
        {\sqrt{1 - (\bar{\mathbf{x}} \cdot \mathbf{w})^{2}}}
        \left[
        	\mathbf{w} - \bar{\mathbf{x}} (\bar{\mathbf{x}} \cdot \mathbf{w})
        \right]
        \mbox{.}
\end{equation*}

The columns of $W$ are given by $\mathbf{w}_{i} = \MyExp{\bar{\mathbf{x}}}{\mathbf{v}_{i}}$.
Once a $W$ is constructed with $S(W) \leq \epsilon$ the optimal $H$ can be found using nonnegative least squares (NNLS). 
So $H$ can be viewed as a function of $W$.
The $W$ itself can be optimized using a direct search method over \TSnm{\bar{\mathbf{x}}} treating $H$, the objective function and the constraints as black-boxes \cite{misc:dreisigmeyer-2006}.
A direct search method is required since $S(W)$ is only Lipschitz continuous \cite{art:kolda-2003}.
This is very similar to the alternating NNLS method except we do not have $W \geq 0$ \cite{art:boutsidis-2009, art:kim-2008, art:lin-2007, art:park-2008}.
We have simply replaced finding $W$ using NNLS with an inequality constrained optimization over \Sn{n-1}.

In \cite{misc:dreisigmeyer-2017} the probabilistic descent method of \cite{art:gratton-2015} was extended to optimization problems on manifolds like \Sn{n-1}.
We show how to specialize the general direct search method in \cite{misc:dreisigmeyer-2017} to the tSNMF.
Only convergence to a local minimum is guaranteed.
Begin by setting
\begin{itemize}
\item $i_{max}$ as the number of iterations to run the algorithm and initialize $i = 0$,
\item $\alpha_{max} = 1$ as the maximum step size,
\item $\alpha_{0} = \alpha_{max}$ as the initial step size,
\item $\theta = 1/2$ as the step size decrease,
\item $\gamma = 2$ as the step size increase, and
\item $\rho(\alpha) = 10^{-3} \alpha^{2}$ as the forcing function.
\end{itemize}
Let $\widehat{X}$ be the data set $X$ with columns normalized to unit length.
To construct the initial $W_{0}$ we will choose columns of $\widehat{X}$ that are `far away' from any previously selected data points.
Set $\mathbf{w}_{0} = \bar{\mathbf{x}}$, the Karcher mean of $\widehat{X}$.  
For $i = 1, \ldots, k$ solve
\begin{equation*}
\mathbf{w}_{i} = 
	\mathrm{arg }
    \max_{\hat{\mathbf{x}} \in \widehat{X}}
    \min_{j = 0}^{i - 1} 
    \arccos( \mathbf{w}_{j} \cdot \hat{\mathbf{x}}) \mbox{.}
\end{equation*}
Let $W_{0} = [ \mathbf{w}_{1} | \cdots | \mathbf{w}_{k} ]$ and $V_{0}$ be the matrix of tangent vectors associated with the columns of $W_{0}$.
Until $S(W_{0}) \leq \epsilon$ iteratively contract the tangent space by a factor of $0.99$ and remap the contracted $V_{0}$ to $W_{0}$. 
Then $H_{0}$ is found by solving the NNLS problem
\begin{equation*}
H_{0} = \mathrm{arg }\min_{0 \leq H} \|X - W_{0}H\|_{F}^{2} 
	\mbox{ where } 
\varepsilon_{0} = \|X - W_{0}H_{0} \|_{F}^{2} \mbox{.}
\end{equation*}
(Note that this method of providing a $W_{0}$ and $H_{0}$ can be replaced with any other method that may be more suitable for a given application.)
Now do the following:
\begin{description}
	\item [Step 1] Set $i \leftarrow i + 1$.
    \item [Step 2 (Optional Search Steps)] Let $D = I_{k} + \mathrm{diag}([d_{1}, \cdots, d_{k}]) $, where the $0 \leq d_{i} \leq \alpha_{i - 1}$ are random numbers, and $Q \in \Rn{k \times k}$ be a random orthogonal matrix.
    \begin{description} 
    	\item [Step 2a (optimal solution with contraction)] Set $W^{\prime} = X H^{\dagger}_{i - 1}$, $H^{\dagger}_{i - 1}$ the pseudo-inverse of $H_{i - 1}$, and normalize the columns to unit length.  
    Let $V^{\prime}$ be the matrix of tangent vectors associated with the columns of $W^{\prime}$. 
    Until $S(W^{\prime}) \leq \epsilon$ iteratively contract the tangent space by a factor of $0.99$ and remap the contracted $V^{\prime}$ to $W^{\prime}$.  
    Then find $H^{\prime}$ by solving the NNLS problem.  
    If $\varepsilon_{i-1} - \varepsilon^{\prime} > \rho(\alpha_{i-1})$ set $V_{i} \leftarrow V^{\prime}$, $\varepsilon_{i} \leftarrow \varepsilon^{\prime}$, $W_{i} \leftarrow W^{\prime}$, $H_{i} \leftarrow H^{\prime}$, $\alpha_{i} = \min(\alpha_{max}, \gamma \alpha_{i-1})$ and Goto \textbf{Step 5}.
    	\item [Step 2b (dilation)] Let $V^{\prime} = V_{i-1} Q D Q^{T}$ and form $W^{\prime}$.
        If $S(W^{\prime}) \leq \epsilon$ then find $H^{\prime}$ and $\varepsilon^{\prime}$ by solving the NNLS problem.
If $\varepsilon_{i-1} - \varepsilon^{\prime} > \rho(\alpha_{i-1})$ set $V_{i} \leftarrow V^{\prime}$, $\varepsilon_{i} \leftarrow \varepsilon^{\prime}$, $W_{i} \leftarrow W^{\prime}$, $H_{i} \leftarrow H^{\prime}$, $\alpha_{i} = \min(\alpha_{max}, \gamma \alpha_{i-1})$ and Goto \textbf{Step 5}.
	\end{description}
	\item [Step 3 (Poll Steps)]  Let $C = \mathrm{diag}([c_{1}, \cdots, c_{k}]) $, $D = I_{k} + \mathrm{diag}([d_{1}, \cdots, d_{k}]) $, $0 \leq c_{i} \leq \alpha_{i - 1}$ and $0 \leq d_{i} \leq \alpha_{i - 1}$ random numbers, $A \in \Rn{n \times (n - k)}$ an orthogonal matrix where $ V_{i-1}^{T} A = 0$, and $Q, U \in \Rn{k \times k}$ and $Z \in \Rn{(n - k) \times k}$ random orthogonal matrices.  
	\begin{description}
		\item [Step 3a] Let $V^{\prime} = V_{i - 1} Q D Q^{T} + A Z C U^{T}$ and form $W^{\prime}$.
        If $S(W^{\prime}) \leq \epsilon$ then find $H^{\prime}$ and $\varepsilon^{\prime}$ by solving the NNLS problem.
If $\varepsilon_{i-1} - \varepsilon^{\prime} > \rho(\alpha_{i-1})$ set $V_{i} \leftarrow V^{\prime}$, $\varepsilon_{i} \leftarrow \varepsilon^{\prime}$, $W_{i} \leftarrow W^{\prime}$, $H_{i} \leftarrow H^{\prime}$, $\alpha_{i} = \min(\alpha_{max}, \gamma \alpha_{i-1})$ and Goto \textbf{Step 5}.
		\item [Step 3b] Let $V^{\prime} = V_{i - 1} Q D Q^{T} - A  Z C U^{T}$ and form $W^{\prime}$.
        If $S(W^{\prime}) \leq \epsilon$ then find $H^{\prime}$ and $\varepsilon^{\prime}$ by solving the NNLS problem.
If $\varepsilon_{i-1} - \varepsilon^{\prime} > \rho(\alpha_{i-1})$ set $V_{i} \leftarrow V^{\prime}$, $\varepsilon_{i} \leftarrow \varepsilon^{\prime}$, $W_{i} \leftarrow W^{\prime}$, $H_{i} \leftarrow H^{\prime}$, $\alpha_{i} = \min(\alpha_{max}, \gamma \alpha_{i-1})$ and Goto \textbf{Step 5}.
	\end{description}
	\item [Step 4] Set $V_{i} \leftarrow V_{i-1}$, $\varepsilon_{i} \leftarrow \varepsilon_{i - 1}$, $W_{i} \leftarrow W_{i-1}$, $H_{i} \leftarrow H_{i-1}$ and $\alpha_{i} = \theta \alpha_{i-1}$.
	\item [Step 5] If $i = i_{max}$ return $W_{i}$ and $H_{i}$, otherwise Goto \textbf{Step 1}.
\end{description}

\section{\label{sec:examples}Example}
The application type we have in mind for the tSNMF is when a high-dimensional dataset can be expressed, at least approximately, as the conical combinations of a few templates.
These applications may differ from using the NMF primarily for clustering, AA where the number of templates is often larger than the embedding dimension, or extensions of NMF that seek convex combinations of a few templates.
They could, however, be similar to problems the singular value decomposition would typically be used for.
That said, the method could still be useful for clustering.

\begin{figure}[!htb]
  \center{\includegraphics[width=\textwidth]
  	{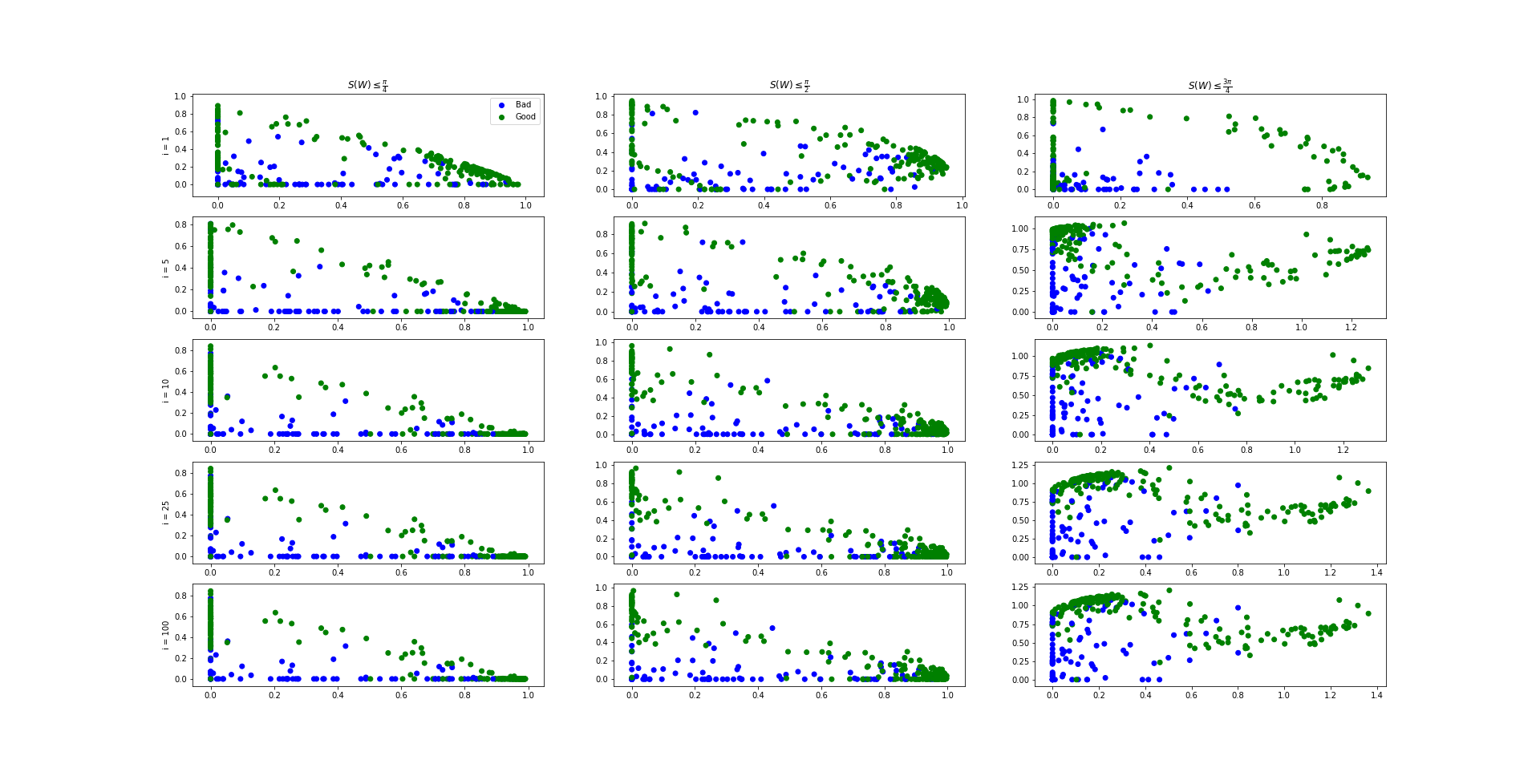}}
  \caption{\label{fig:ionosphere}UCI Machine Learning Repository ionosphere data set example.  The reduction was onto two templates ($k = 2$).  The plots are of the entries of $H$ for $\epsilon = \frac{\pi}{4}, \frac{\pi}{2}, \mbox{ and } \frac{3\pi}{4}$ with $i_{max} = 1, 5, 10, 25, \mbox{ and } 100$.  The x-axis (y-axis) corresponds to the entries in the first (second) row.}
\end{figure}

Here we look at reducing the dimensionality of the Ionosphere Data Set from the UCI Machine Learning Repository \cite{art:sigillito-1989}.
The data set has both positive and negative attribute values.
Each data point is assigned a class of either `b' (bad) or `g' (good).
We did some preprocessing of the data before our experiment.
In the first column a value of 0 always corresponded with a classification of `b' in the final column so all of those rows were removed.
Now the first and second columns of the data set had a single value (1 and 0 respectively) and were removed.
The final preprocessed data points all have unit 2-norm length.
The results are shown in Figure~\ref{fig:ionosphere}.

With $\epsilon = \frac{\pi}{4}$ one of the templates roughly corresponds to the `bad' data points while the other is a `good' data point template.
Along the first template there are also many `good' data points with the `bad' data points tending to be closer to the origin versus the `good' data points.
As $\epsilon$ increases the templates begin to characterize two extreme `bad' data templates.
The remaining data points are combinations of these two templates with the `good' data points generally being further from the origin versus the remaining points in the `bad' class.

\section{\label{sec:discussion}Discussion}
We've developed the tight semi-nonnegative matrix factorization where a data set $X$ is modeled as conical combinations of unit length templates given by the columns of a matrix $W$.
We make this approximation tight by requiring that the maximum geodesic distance $S(W)$ over the unit hypersphere \Sn{n-1} be minimized as well as the approximation error.
By minimizing $S(W)$ the original data $X$ can be modeled but as little else as possible.
As a multi-objective optimization, the solution is given by a Pareto frontier where there is a trade-off between 1) being able to model the data $X$ and 2) including extraneous volume in the cone defined by $W$.

The tight semi-nonnegative matrix factorization is similar to the convex matrix factorization \cite{art:ding-2010} and archetypal analysis \cite{art:cutler-1994}.
It differs from both by not requiring that the templates be convex combinations of the original data.
It also differs from the method in \cite{misc:javadi-2017} by considering $S(W)$ instead of the distance of the columns of $W$ from the convex hull of the original data.

\bibliographystyle{siamplain}
\bibliography{references}
\end{document}